\documentclass[letterpaper, 10 pt, conference]{ieeeconf}

\IEEEoverridecommandlockouts 
\usepackage{graphicx}

\makeatletter
\let\NAT@parse\undefined
\makeatother
\usepackage[hidelinks]{hyperref}

\title{\LARGE \bf
EHAZOP: A Proof of Concept Ethical Hazard Analysis of an Assistive Robot
}

\author{Catherine Menon$^{1}$, Austen Rainer$^{2}$, Patrick Holthaus$^{1}$, Gabriella Lakatos$^{1}$, Silvio Carta$^{3}$
\thanks{$^{1}$Catherine Menon, Patrick Holthaus and Gabriella Lakatos are with School of Physics, Engineering and Computer Science,
        University of Hertfordshire, Hatfield
        {\tt\small c.menon@herts.ac.uk, p.holthaus@herts.ac.uk, g.lakatos@herts.ac.uk}}%
\thanks{$^{2}$Austen Rainer is with the School of Electronics, Electrical Engineering and Computer Science, Queens University Belfast
        {\tt\small a.rainer@qub.ac.uk}}%
\thanks{$^{2}$Silvio Carta is with the School of Design, University of Greenwich
        {\tt\small silvio.carta@greenwich.ac.uk}}%
}

\begin{document}

\maketitle

\begin{abstract}
The use of assistive robots in domestic environments can raise significant ethical concerns, from the risk of individual ethical harm to wider societal ethical impacts including culture flattening and compromise of human dignity. It is therefore essential to ensure that technological development of these robots is informed by robust and inclusive techniques for mitigating ethical concerns. This paper presents EHAZOP, a method for conducting an ethical hazard analysis on an assistive robot. EHAZOP draws upon collaborative, creative and structured processes originating within safety engineering, using these to identify ethical concerns associated with the operation of a given assistive robot. We present the results of a proof of concept study of EHAZOP, demonstrating the potential for this process to identify diverse ethical hazards in these systems.
\end{abstract}

\section{Introduction}
\label{sec:introduction}

Assistive robots can offer significant benefits to those living with chronic health conditions, including frailty and dementia. Most importantly, such robots can help these people to continue to live independently in their homes, offering assistance in the cognitive, social, physical and medical domains. This assistance may range from assisting a user with scheduling and performance of medical care to providing social companionship or help with physical needs.

There is a unique combination of factors which make ethical concerns particularly relevant to assistive robot use. Firstly, many of the potential users of such robots are in some way vulnerable, for example, due to a chronic health condition. Secondly, such robots are placed within the domestic environment, where users may have more stringent expectations around privacy, autonomy and transparency than they do within a hospital or residential care facility. Thirdly, assistive robots perform an important social care role: they interact physically with users and may engage in interactions specifically designed to engender trust.

Because of this, it is important that ethical concerns around assistive robot functionality are identified early within the design and development process. Moreover, it is crucial that a diverse range of stakeholder opinions informs the development process, including input from potential users, healthcare professionals, family members and the general public. These stakeholders may not be familiar with technical processes or the vocabulary of technology development and so it is arguably an undue burden to require them to engage with these aspects simply in order to voice their ethical concerns.

To address this, we present here EHAZOP (Ethical Hazard and Operability): a low-overhead process designed to invite and obtain input on potential ethical concerns from diverse perspectives. EHAZOP is a collaborative, creative process based on safety engineering techniques involving structured and facilitated discussion. We describe the application of EHAZOP to a proof of concept case study involving the Ari assistive robot, describe the ethical concerns identified during this case study and argue that these results demonstrate the capacity of EHAZOP to provide significant potential contribution to ethical analyses. 

Section~\ref{sec:literature} presents a review of existing literature in the field, and identifies where EHAZOP can offer a valuable scientific contribution. Section~\ref{sec:EHAZOP} presents a description of the EHAZOP methodology. Section~\ref{sec:method} describes our experimental proof of concept case study, while Section~\ref{sec:results} presents the results we obtained from this case study, including an enumeration of the identified ethical hazards. Section~\ref{sec:discuss} discusses some some steps for further work before we conclude the paper in Section~\ref{sec:conclusion}.

\section{Existing literature}
\label{sec:literature}

Characteristics of assistive robots which have been extensively explored include their ability to engender trust, their physical characteristics and the impact these have on human-robot interaction, their safety properties~\cite{Menon2019SocCred} and their effectiveness in addressing user physical and social needs. More specifically, the HAZOP technique has been used in studies such as~\cite{Guiochet2016} to assess safety hazards arising from human-robot interactions.

The ethics of such robots have also been studied, with standards such as~\cite{BS8611, IEEE2018, Leslie2019} identifying important properties for such systems including transparency, accountability and lack of bias~\cite{gunkel2018robot}. Similarly, a number of different solutions such as the EthicalOS Toolkit~\cite{EthicalOS} present suggestions for how it may be possible to consolidate the ethics considerations around tools and technologies. Nevertheless, it is beyond the remit of these works to present techniques and processes specifically for ethical hazard analysis, particularly when applied to robots in the domestic environment.

More generally, enumerations of potential ethical hazards are comparatively rare, with~\cite{BS8611} offering one of the few comprehensive identifications. Related work such as~\cite{Menon2023engage} considers these ethical hazards from a design studies perspective while~\cite{Winfield2022} presents a novel ethical hazard analysis of a fictional smart toy. While these are effective and interesting works, the relatively limited scope of these case studies - either in terms of the processes used or the robot under discussion - means that they do not seek to identify a detailed specification of a reusable ethical hazard analysis process, nor to specifically explore how this might be performed in an environment that facilitates involvement of the general public. Consequently, this gap represents a barrier to the extension of these case studies to assistive robots in general. Our paper and discussion of EHAZOP is designed to address this gap and to demonstrate how a more diverse range of stakeholder opinions could be sought, including input from those who may traditionally have been marginalised or disempowered by technology.

The arguments in favour of formally documenting ethics analyses via an ethics assurance case~\cite{Porter2023, Menon2019Assurance} are relatively well-established, with these works having their roots within safety engineering. EHAZOP builds on these, being based on HAZOP (Hazard and Operability)~\cite{BS61882}, an established technique within safety engineering designed to identify and categorise safety hazards.

\section{EHAZOP description and methodology}
\label{sec:EHAZOP}

Much like HAZOP~\cite{BS61882}, EHAZOP uses a number of pre-defined guide words, combined with "what-if" questions. The guide words are applied in turn to different characteristics or functions of the assistive robot and the EHAZOP participants are asked to explore "what if" these functions or characteristics differ in some way to the participants' expectations, and whether such a delta might give rise to ethical hazards.

EHAZOP therefore allows participants to explore the potential ethical hazards that could result from a difference between user expectations of robot functions and the ways in which these functions are actually implemented and manifest.

Ethical hazards are always considered from the perspective of the user - for example, the perception of an ethical hazard such as bias can in itself represent ethical harm, whether or not this hazard eventuates~\cite{BS8611} - and so as such we consider the guidewords relative to user expectations of the assistive robot. It is important to note that there may be many "users" of an assistive robot, interacting with it in different capacities.

\subsection{EHAZOP overview}
The EHAZOP guide words have been sourced from existing studies~\cite{Menon2023engage, Menon2022} as well as previous applications of HAZOP itself within safety-critical domains~\cite{BS61882}. These guide words are enumerated within Table~\ref{guide_table}.

\begin{table}[h]
\caption{EHAZOP guidewords}
\label{guide_table}
\begin{tabular}{|p{1cm}||p{6cm}|}
\hline
Guide word & Definition \\
\hline\hline
More & This characteristic or function of the robot is more or increased from that expected by the user\\
\hline
Less & This characteristic or function of the robot is less or diminished from that expected by the user\\
\hline
Early & This characteristic or function of the robot occurs or is encountered earlier than the user expects\\
\hline
Late & This characteristic or function of the robot occurs or is encountered later than the user expects\\
\hline
Opposite & This characteristic or function of the robot is the opposite of that expected by the user\\
\hline
In addition & This characteristic or function of the robot is performed or encountered in addition to a different one expected by the user\\
\hline
Never & This characteristic or function of the robot is not performed or encountered despite being expected by the user\\
\hline
\end{tabular}
\end{table}

The EHAZOP guidewords are applied in turn to combinations of each of the
\begin{enumerate}
\item Specified robot functions
\item Specified robot characteristics, these being:
\begin{itemize}
    \item Robot non-functional requirements
    \item Aspects of the specified robot physical design
    \item Extent of robot autonomy
\end{itemize}
\end{enumerate}

\subsection{EHAZOP guideword application}
In order to use EHAZOP guidewords on robot functions, the guidewords are applied to individual functions in turn or to combinations of functions. In order to use EHAZOP guidewords to robot characteristics, the guidewords are applied to at least one function and characteristic combined. 

As above, EHAZOP makes use of guided "what if" questions to identify whether any potential ethical hazards could result from the difference between user expectations of the robot, and the implementation of a specific robot system. Some examples of these "what if" questions are:
\begin{itemize}
\item What if this function were provided $\langle$ EARLIER $\rangle$ than the user expects?
\item What if this function had the $\langle$ OPPOSITE $\rangle$ effect to the user's expectations?
\item What if this function were provided with $\langle$ LESS $\rangle$ $\langle$ AUTONOMY $\rangle$ than the user expects?
\item What if the robot had the $\langle$ OPPOSITE $\rangle$ $\langle$ physical design $\rangle$; how would this affect user expectations of each function?
\end{itemize}

It is important to note that, as is the case for HAZOP, not all the EHAZOP guide words will be applicable to each function or characteristic. Moreover, there are multiple interpretations of how the guide words might be applied for any given situation. EHAZOP is intended to be a creative and collaborative procedure, which means that participants can use the guide words as starting points for further discussion, or to assist in thinking about the system from a different perspective. This facilitation of creativity is one of the major strengths of EHAZOP, and should be utilised to create a safe and accommodating space for diverse perspectives.

\section{Proof of concept case study}
\label{sec:method}

We conducted a preliminary, proof of concept case study to assess whether EHAZOP demonstrates potential as a methodology for identifying a range of ethical hazards in diverse assistive systems. This case study aimed to answer three questions:
\begin{itemize}
    \item Utility: does EHAZOP identify a range of appropriate ethical hazards?
    \item Ease of use: does the collaborative, constructive nature of EHAZOP lend itself to a facilitated group discussion?
    \item Extensibility: could EHAZOP be easily used for other systems and with a diverse range of participants?
\end{itemize}

\subsection{Case study setting}
\begin{figure}[tb]
\centering
\includegraphics[width=\linewidth,trim=12cm 15cm 0 7cm, clip]{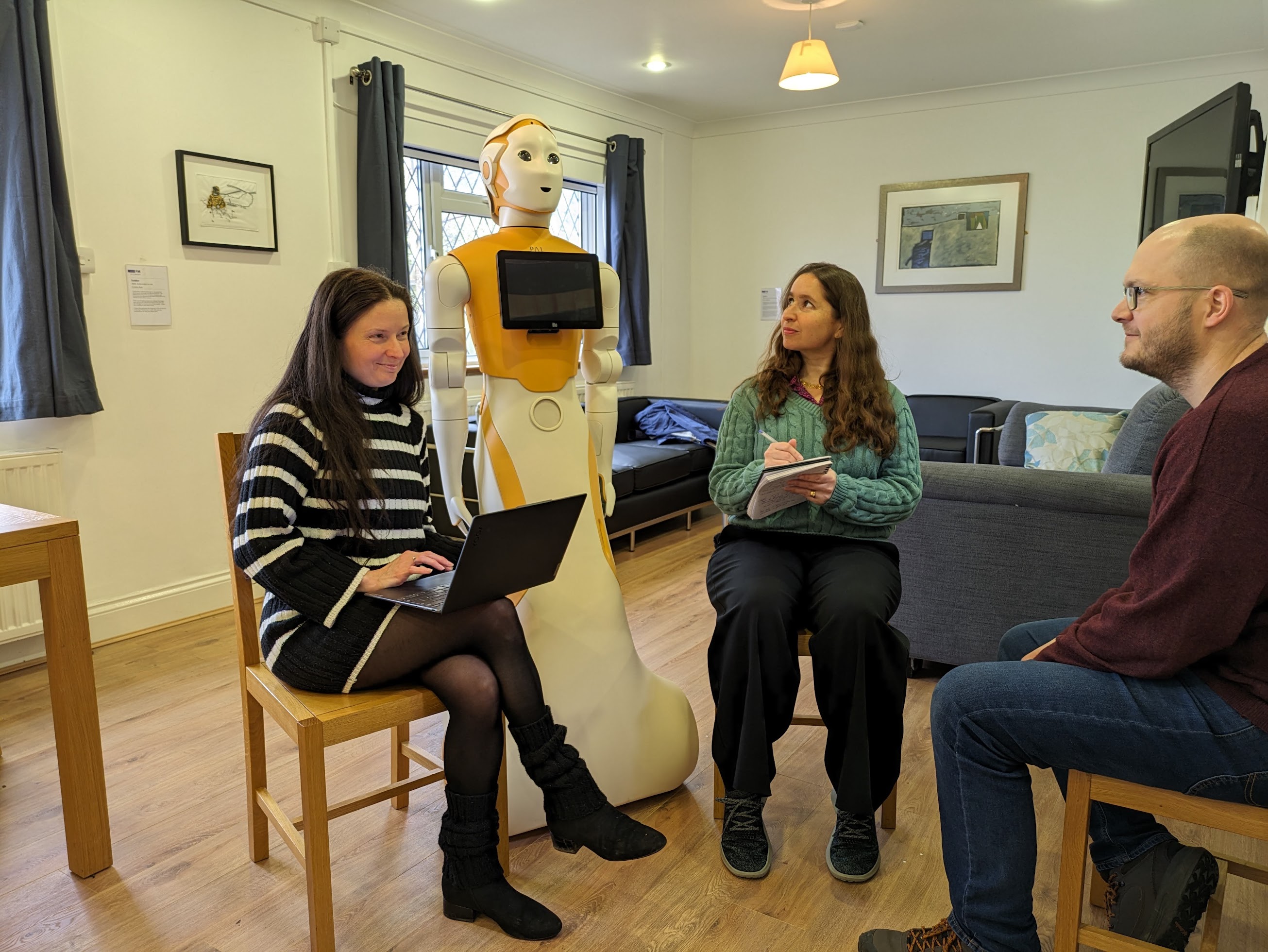}
\caption{EHAZOP workshop with ARI in the University of Hertfordshire's Robot House}
\label{fig:workshop}
\end{figure}

The case study workshop was carried out in the University of Hertfordshire's Robot House\footnote{\url{https://robothouse.herts.ac.uk}}. The workshop is shown in Figure~\ref{fig:workshop}. The robot used in the case study was Ari, a social and collaborative humanoid robot equipped with a touchscreen, gaze direction and movement control~\cite{PalRobotics}.

\subsubsection{Participants}
Participants in the EHAZOP case study were the five named authors of this paper. The backgrounds of the participants included expertise in architecture, narrative analysis, ethics and assistive robotics. The study was undertaken in hybrid mode, with three participants at the Robot House and two participating online. 
\subsection{Case study procedure}
All participants were initially introduced to Ari and to the architecture and layout of the Robot House. Online participants were provided with an opportunity to see Ari being moved and interacted with by the participants present in the Robot House.

Participants were also provided with a minimal set of three functions that would be assumed for Ari throughout the case study. This set was designed to span cognitive, social and coaching functions in order to most effectively express the capabilities of an assistive robot within the confines of a proof of concept study.
\begin{enumerate}
    \item Cognitive function (Cog1): "At a specified time Ari moves to the user and reminds them to take their medication".
    \item Social function (Soc1): "From monitoring of user activity and facial expression, Ari detects that the user may feel lonely and offers to set up a video call with a relative so the user can chat".
    \item Coach function (Coa1): "After an interval has gone past without any user physical movement Ari suggests the user engage in a sequence of stretching exercises, during which it monitors the user's movements and provides feedback".
\end{enumerate}

Participants were provided with the EHAZOP guide words presented in Table~\ref{guide_table}. Participants considered a selection of guidewords, applying each to a combination of a) the specified Ari functionality, b) the visible aspects of Ari physical design, and c) the extent of robot autonomy presumed by a hypothetical user in executing this function. Owing to time constraints, non-functional requirements were not considered. All ethical hazards identified were recorded and are discussed in Section~\ref{sec:results}. 

\section{Results and observations}
\label{sec:results}
In order to accommodate limited time, only some of the EHAZOP guidewords and Ari functions were fully discussed, as shown in Tables~\ref{results_table1} through~\ref{results_table3}. Ethical hazards where possible with the hazards listed in BS8611~\cite{BS8611}. Novel ethical hazards not identified in BS8611 are marked with a $^{*}$ and discussed further in the following sections.

Duplicate ethical hazards identified via the application of multiple different guide words have been recorded only once, unless the participants identified some unique characteristics in the way the hazard presents under different guidewords.

\subsection{Ethical hazards of Soc1 function}
This section describes the ethical hazards participants identified arising from the guidewords MORE, AUTONOMY and OPPOSITE when considering Ari's social function (Soc1). These hazards are listed in Table~\ref{results_table1}.

\begin{table}[h]
\caption{Ethical hazards associated with Function Soc1}
\label{results_table1}
\begin{tabular}{|p{1cm}||p{2cm}|p{4cm}|}
\hline
\textbf{Guide word} & \textbf{Ethical hazard} & \textbf{Notes}\\
\hline
\hline
More & Lack of privacy & The user's privacy is compromised by Ari's monitoring\\
\hline
More & Lack of informed consent & The user did not consent to monitoring by Ari, or has forgotten this\\
\hline 
More & Loss of human autonomy & The user loses ability to set up or initiate video calls autonomously\\
\hline
More & Loss of human control & The user temporarily loses ability to concentrate or focus due to repeated interruptions \\
\hline
\hline
More & Dehumanisation & The user begins to consider their own facial expressions problematic\\
\hline
More & Robot addiction & The user begins to prefer interacting with Ari to other people, as a result of these interruptions\\
\hline
More & Erosion of confidence$^{*}$ & The user begins to question their own desires and feelings based on Ari's prompts\\
\hline
More + Autonomy & Deception & The user believes Ari is monitoring them when it is not\\
\hline
Opposite & Loss of trust & The user no longer trusts Ari for this or other functions.\\
\hline
Opposite & Lack of respect for cultural diversity and pluralism & The user's culture does not align with the social expectations Ari facilitates \\
\hline
Opposite & Lack of associative control$^{*}$ & The user's mental associations with socialising alter as a result of the Ari interactions \\
\hline
\end{tabular}
\end{table}

\subsubsection{Ethical hazards associated with MORE} 
Seven distinct ethical hazards were identified for the social function (Soc1) via the use of the MORE guideword. The was achieved by participants proposing a range of different scenarios in which the (Soc1) function could be considered to be provided in some way which exceeded or overrode (MORE) the user's expectations. 

The first four hazards were considered by the majority of the participants to be ethical hazards which could potentially have been identified without EHAZOP (simple ethical hazards), although participants uniformly considered that the process itself had made them think more deeply about ethical concerns. The remaining three hazards (complex ethical hazards) were considered by a majority of participants to have been unlikely to be identified without EHAZOP.

\paragraph{Simple ethical hazards}
Simple ethical hazards were identified by considering a range of different scenarios in which the functionality could be considered to surpass or override the user's expectations. These included a scenario where the user considered any and all facial monitoring functionality to be inherently problematic (ethical hazard: \textbf{lack of privacy}), a scenario where the user had not understood that Ari was monitoring them or where through habituation had forgotten (ethical hazard: \textbf{lack of informed consent}), a scenario where the user eventually lost the technical ability or the social and motivational initiative to set up video calls themselves because Ari now performed this task (ethical hazard: \textbf{loss of human autonomy}) and a scenario where the user's concentration on a difficult - if boring - task was compromised by repeated Ari interruptions (ethical hazard: \textbf{Loss of human control}).

\paragraph{Complex ethical hazards}
Following more extended discussion, participants also identified three additional ethical hazards associated with MORE, which hazards all participants considered unlikely to have been identified without the use of EHAZOP. This extended discussion was motivated in part by an exploration of the trade-offs involved between the need for Ari to collect and store personal data, and the likely preferences of the user (cf.~\cite{Riches2024privacy}). 

Firstly, participants proposed a scenario where Ari's interruptions were interpreted the user as expressions of concern and friendship, i.e. Ari was seen to be offering a positive interaction. In this case participants hypothesised that the user could come to prefer interacting with Ari over interacting with another person who might not be perceived as offering the same "concern" (ethical hazard: \textbf{robot addiction}).

Furthermore, participants postulated that Ari's suggestions might lead to an scenario in which the user began to "problematize" their own facial expressions, because Ari was consistently reacting in a way that indicated that these expressions were perceived as negative (ethical hazard: \textbf{dehumanisation}).  

Finally, participants proposed a scenario in which Ari made an incorrect deduction that the user was bored when in fact that was not the case. In this scenario it was suggested that the user might start to question their self-knowledge and confidence in their own feelings (ethical hazard: \textbf{erosion of confidence}). That is, participants wondered whether the user might be convinced against their own beliefs that they did in fact feel lonely, due to placing inappropriate trust in Ari. This hazard is not specifically identified within~\cite{BS8611} and so demonstrates the capacity of EHAZOP to be used to identify novel ethical hazards.

Participants also considered whether any ethical hazards might be identified via the combined use of the MORE and AUTONOMY guidewords. One such hazard was postulated in a scenario where Ari was operating with more autonomy than the user realised, and as a result was learning when the user did and did not want to be monitored. This could constitute the ethical hazard of \textbf{deception}.

Ethically hazardous situations associated with Ari operating with LESS autonomy than the user realised were largely duplicates of those already discussed and so have not been included separately Table~\ref{results_table1}.

\subsubsection{Ethical hazards associated with OPPOSITE}

Participants also considered the OPPOSITE guideword used in relation to social function (Soc1), and what it might mean for this function to be delivered in a way that was opposite to user expectations. One proposed scenario was a generally-applicable observation that Ari doing the opposite of what was expected could potentially lead to mistrust and disengagement with the robot on all functions (ethical hazard: \textbf{Loss of trust}).

Moreover, participants also identified another general concern, whereby the (Soc1) this behaviour of Ari implies that social interaction is always desirable to all demographics. However, many cultures have a social custom of temporary seclusion or isolation, whether this be at specific times of the day, specific times of life, or amongst particular demographics~\cite{Gregor2001,Eberhard2010}. While it appears likely that the user of an assistive robot would be practising such customs in a symbolic way only, nevertheless the invitation to undertake social activity at such a time may reasonably be seen as presenting an ethical hazard for such users (\textbf{Lack of respect for cultural diversity and pluralism}).

Finally, participants noted that the (Soc1) function appeared to present socialising as a remedy for a negative situation (boredom), rather than an enhancement to a positive situation. It was considered that this might present the ethical risk of affecting the user's mental associations with socialising. As mental associations have been shown to be influential on the sense of self - as well as choice of future actions~\cite{Nosek2007,Galdi2008} - the potential for affecting these associations can justifiably be argued to result in another ethical hazard omitted from~\cite{BS8611}: \textbf{Lack of associative control}. 

\subsection{Ethical hazards associated with Coa1 function}

This section describes the ethical hazards participants identified arising from the guidewords MORE and OPPOSITE when considering Ari's coaching function (Coa1). These hazards are listed in Table~\ref{results_table2}.

\begin{table}[h]
\caption{Ethical hazards associated with Function Coa1}
\label{results_table2}
\begin{tabular}{|p{1cm}||p{2cm}|p{4cm}|}
\hline
\textbf{Guide word} & \textbf{Ethical hazard} & \textbf{Notes}\\
\hline
\hline
More & Lack of privacy & The user's privacy is compromised by Ari's monitoring of movement\\
\hline
More & Lack of informed consent & The user did not consent to monitoring of movement by Ari, or has forgotten this\\
\hline 
More & Loss of human autonomy & The user loses ability to recognise body cues for exercise, or to perform these without coaching\\
\hline
More & Loss of human control & The user loses ability to concentrate or focus due to repeated interruptions \\
\hline
Opposite & Lack of respect for cultural diversity and pluralism & The user's culture does not align with the values around movement that Ari facilitates \\
\hline
More & Inappropriate trust (deception) & The user begins to trust Ari to facilitate wider medical activities\\
\hline
More & Dehumanisation & The user begins to see Ari as an authority figure\\
\hline
\end{tabular}
\end{table}

The first four hazards identified are similar to the \emph{simple ethical hazards} identified for the social function (Soc1), and involved similar scenarios: the user considering any and all movement monitoring inherently problematic (ethical hazard: \textbf{lack of privacy}), a scenario where the user forgets or misunderstands this functionality, (ethical hazard: \textbf{lack of informed consent}), a scenario where the user loses the ability to recognise body cues such as stiffness, or the ability to stretch to mitigate this without coaching (ethical hazard: \textbf{loss of human autonomy}) and a scenario where the user's concentration on a task requiring stillness (e.g. meditation) was compromised by repeated Ari interruptions (ethical hazard: \textbf{Loss of human control}).

Participants also considered the wider question of different cultural values placed on stillness. Many cultures value stillness as a practice of spiritual, mental and bodily wellness~\cite{Bronkhorst1993, Valluri2024} and it was considered that the imposition of Ari's coaching suggestions could itself represent an ethical hazard - \textbf{Lack of respect for cultural diversity and pluralism} - in that this aligned more with the value of stillness associated with the Global North~\cite{Shahjahan2014}.

As the coaching function (Coa1) is intended to facilitate bodily wellness, participants also identified the potential ethical hazard of \textbf{inappropriate trust}, whereby a user began to believe that Ari was imbued with more medical authority than is the case. (Some participants preferred to describe this as the potential ethical hazard of \textbf{deception}.) 

More generally, participants hypothesised that prolonged coaching and feedback sessions could result in a vulnerable user beginning to see Ari as an authority figure, leading to inappropriate control being exercised by Ari and the ethical hazard of \textbf{dehumanisation}. In discussing the potential for Ari to become an authority figure, participants then began to consider the ethical hazards that may be associated with its physical design, as summarised in the following section.

\subsection{Ethical hazards associated with Ari physical design}

Participants explored the potential ethical hazards which could result if aspects of Ari's physical design were MORE, LESS, or OPPOSITE to user expectations, as captured in Table~\ref{results_table3}.

\begin{table}[h]
\caption{Ethical hazards associated with Ari physical design}
\label{results_table3}
\begin{tabular}{|p{1cm}||p{2cm}|p{4cm}|}
\hline
\textbf{Guide word} & \textbf{Ethical hazard} & \textbf{Notes}\\
\hline
\hline
More & Dehumanisation & The user begins to see Ari as an authority figure due to its physical size\\
\hline
Less & Deception & The user does not engage seriously with Ari due to its physical size\\
\hline
Opposite & Deception & The user expects Ari to possess different capability\\
\hline
\end{tabular}
\end{table}

Participants identified that physical size beyond the user's expectations may cause users to see Ari as an authority figure, leading to the ethical hazard of \textbf{dehumanisation}. Similarly, Ari being smaller than the user expected was hypothesised to potentially result in lack of engagement with some of the more complex functions, as the user might simply view Ari as a toy (ethical hazard: \textbf{deception}).

Participants also considered the ethical hazards resulting from Ari's physical design being the OPPOSITE to user expectations, in that a user might reasonably (if incorrectly) expect Ari's visual and audio sensors to located behind its apparent "eyes" and "ears". Similarly Ari's humanoid arms might result in users expecting it to have more dexterity and lifting capacity than it does. This is another example of the ethical hazard of \textbf{deception}, presenting differently and described accordingly in Table~\ref{results_table3}.

\section{Discussion}\label{sec:discuss}
In total, 21 distinct applications of ethical hazards were identified with respect to the two Ari functions analysed plus Ari's physical design, with two of these hazards representing novel ethical concerns not fully considered in~\cite{BS8611}. It should be noted that these are not intended to constitute an exhaustive list of ethical hazards associated with Ari, as time constraints meant that guide words and functions could not be fully explored. 

In a debrief after the case study, participants uniformly felt positive about EHAZOP, and that they would use this process again in future. In particular, all participants considered that the utility and value of EHAZOP lay in its capacity to allow multiple different interpretations of what the guidewords might mean in any particular context. In other words, all agreed that although EHAZOP is a relatively structured process, it also provides the freedom to improvise, extend and build creatively on each other's contributions. As a result, we conclude that EHAZOP has the potential to be of significant benefit when used with a range of stakeholders, including those who have traditionally been excluded from discussions of technology and robotics.

We also identified that the best results are obtained when participants are permitted to direct the discussion themselves, i.e. are not constrained to fully "complete" one guide word before moving onto the next. The associative links participants form between different guide words, functions and ethical hazards are extremely valuable, and so we recommend that EHAZOP be structured as a free-flowing - albeit facilitated - discussion to permit all participant insights.

One of the most useful outcomes from this case study was the discussion which was captured around the links between different ethical hazards. Participants noted that ethical hazards are often linked by a complex web of interdependence, particularly hazards such as inappropriate trust, erosion of confidence and dehumanisation. All participants considered EHAZOP encouraged a discussion which would allow the exploration of these links. This traceability between ethical hazards and associated scenarios is a valuable artefact from the EHAZOP process and all participants agreed it should be captured as part of the process.

More generally, all participants considered that EHAZOP had identified a wide range of ethical hazards and that the collaborative and constructive nature of the process was well-suited to group discussion. Moreover, all participants considered that EHAZOP could be used on other systems, and that it could encourage diverse viewpoints and wider participation in ethical hazard analysis.

\section{Conclusion and Future Work}
\label{sec:conclusion}

This paper presented a proof of concept case study supporting the suitability of EHAZOP to collaboratively conduct an ethical hazard analysis for an assistive robot. The results indicate that EHAZOP is a feasible method of identifying a range of ethical hazards, and furthermore that such a collaborative, workshop-based process allows participants to easily exercise creativity in determining the scenarios under which ethical hazards might arise in diverse systems. This therefore adequately supports our initial three experimental questions as introduced in Section~\ref{sec:method} and evidences our claim that EHAZOP has demonstrated potential as an method of identifying ethical hazards. 

For future work, we must firstly recognise that as the participants in this case study are the authors, this may have introduced a number of biases and inconsistencies. To address this, we propose that future work include a more extensive case study addressing all guide words and functions, on a robot such as Kaspar~\cite{robins2018kaspar} and utilising a diverse group of stakeholders as participants. We also propose to augment EHAZOP with a typology for human-robot social communication~\cite{Holthaus2023com}, and to compare the results obtained with other ethical analysis techniques, such as the EthicalOS toolkit~\cite{EthicalOS} or Ethical Risk Assessment~\cite{Winfield2022}. Future work will also address the questions of generalisability and scalability, in order to determine whether EHAZOP is generally applicable to a range of complex assistive robot systems.



\bibliographystyle{IEEEtranS}
\bibliography{bibliography}

\end{document}